\documentclass[10pt,journal,compsoc]{IEEEtran}

\ifCLASSOPTIONcompsoc
  \usepackage[nocompress]{cite}
\else
  \usepackage{cite}
\fi

\ifCLASSINFOpdf
\else
\fi

\usepackage{dblfloatfix}
\usepackage{booktabs}
\usepackage{graphicx}

\usepackage{color}

\usepackage{xcolor}

\usepackage{makecell}

\usepackage[hyphens]{url}
\usepackage[hidelinks, pagebackref=true,breaklinks=true,colorlinks,bookmarks=false]{hyperref}
\hypersetup{breaklinks=true}
\begin{document}
%
\title{VIPriors 2: Visual Inductive Priors for\\Data-Efficient Deep Learning Challenges}
%
%
%
%

\author{Attila~Lengyel,
        Robert-Jan~Bruintjes,
        Marcos~Baptista~Rios,
        Osman~Semih~Kayhan,
        Davide~Zambrano,
        Nergis~Tomen
        and~Jan~van~Gemert
\IEEEcompsocitemizethanks{
\IEEEcompsocthanksitem A. Lengyel, R.J. Bruintjes, N. Tomen and J. van Gemert are with Delft University of Technology.\protect\\
E-mail: a.lengyel@tudelft.nl
\IEEEcompsocthanksitem M. Baptista Rios is with Gradiant.
\IEEEcompsocthanksitem O. S. Kayhan is with Bosch Security Systems B.V.
\IEEEcompsocthanksitem D. Zambrano is with Synergy Sports.
}
}

%
%

\IEEEtitleabstractindextext{%
\begin{abstract}

The second edition of the "VIPriors: Visual Inductive Priors for Data-Efficient Deep Learning" challenges featured five data-impaired challenges, where models are trained from scratch on a reduced number of training samples for various key computer vision tasks. To encourage new and creative ideas on incorporating relevant inductive biases to improve the data efficiency of deep learning models, we prohibited the use of pre-trained checkpoints and other transfer learning techniques. The provided baselines are outperformed by a large margin in all five challenges, mainly thanks to extensive data augmentation policies, model ensembling, and data efficient network architectures.
\end{abstract}

\begin{IEEEkeywords}
Visual inductive priors, challenge, image classification, object detection, instance segmentation, action recognition.
\end{IEEEkeywords}}

\maketitle

\IEEEdisplaynontitleabstractindextext

%
\IEEEpeerreviewmaketitle


%
%
%
%

\IEEEraisesectionheading{\section{Introduction}\label{sec:introduction}}

\IEEEPARstart{D}{ata} is fueling deep learning. Data is costly to gather and expensive to annotate. Training on massive datasets has a huge energy consumption adding to our carbon footprint. In addition, there are only a select few deep learning behemoths which have billions of data points and thousands of expensive deep learning hardware GPUs at their disposal. The Visual Inductive Priors for Data-Efficient Deep Learning workshop (VIPriors) aims beyond the few very large companies to the long tail of smaller companies and universities with smaller datasets and smaller hardware clusters. We focus on data efficiency through visual inductive priors.

For the second year running we organized the Visual Inductive Priors for Data-Efficient Deep Learning workshop. The 2021 edition of this workshop took place at ICCV. As part of our workshop, we organize challenges to stimulate research into data-efficient computer vision. For these challenges, the task is to train computer vision models on small subsets of publicly available datasets. 
We challenge the competitors to submit solutions that can learn a good model of the dataset without access to the scale of data that powers state-of-the-art deep computer vision.

In this report, we discuss the outcomes of the second edition of these challenges. We discuss the setup of each challenge and the solutions that achieved the top rankings. We find that the top competitors in all challenges heavily rely on model ensembling and data augmentation to make their solutions data-efficient. However, this year's top submissions also include novel ideas and contributions. To highlight these works, we award a jury prize for each challenge to the most interesting submission. 

\begin{table*}[ph]
\centering
\caption{Overview of challenge submissions. Bold-faced methods are contributions by the competitors.}
\label{tab:conclusion}
\renewcommand{\arraystretch}{1.4}
\scalebox{0.9}{
\begin{tabular}{@{}lllllc@{}}
\toprule
Ranking & Teams & Encoder architectures & Data augmentation & Methods & Main metric \\ \midrule
\multicolumn{2}{@{}l}{\textbf{Classification}} & & & \\ \midrule
1 & \makecell[l]{\textbf{Sun et al.}} & ResNeSt~\cite{zhang2020resnest} & \makecell[l]{AutoAugment~\cite{cubuk2018autoaugment}, MixUp~\cite{zhang2018mixup},\\ CutMix~\cite{yun2019cutmix}} & \makecell[l]{BlurPool~\cite{zhang2019making},\\ stochastic depth~\cite{huang2016deep}} & \textbf{75.5} \\
2 & \makecell[l]{J. Wang et al.} & \makecell[l]{ResNeSt~\cite{zhang2020resnest}, TResNet~\cite{ridnik2021tresnet},\\ RexNet~\cite{han2021rethinking}, RegNet~\cite{radosavovic2020designing},\\ Inception-ResNet~\cite{szegedy2017inception}} & \makecell[l]{AutoAugment~\cite{cubuk2018autoaugment}, MixUp~\cite{zhang2018mixup},\\ CutMix~\cite{yun2019cutmix}, label smoothing~\cite{szegedy2016rethinking}} & \makecell[l]{DSB-Focalloss} & 75.2           \\
2$\,\dagger$ & \makecell[l]{Guo et al.} & \makecell[l]{EfficientNet-b5/b6/b7~\cite{tan2019efficientnet}, \\ DSK-ResNeXt101~\cite{bruintjes2021vipriors, xie2017aggregated},\\ ResNet-152~\cite{he2015deep}, SEResNet-152~\cite{xie2017aggregated}} & \makecell[l]{AutoAugment~\cite{cubuk2018autoaugment}, MixUp~\cite{zhang2018mixup},\\ CutMix~\cite{yun2019cutmix}, label smoothing~\cite{szegedy2016rethinking}, \\ dropout~\cite{srivastava14a}, random erasing~\cite{zhong2020random}} & \makecell[l]{\textbf{Contrastive Regularization},\\ Mean Teacher~\cite{tarvainen2017mean},\\ Symmetric Cross Entropy~\cite{wang2019symmetric}} & 74.3           \\
3 \& J & \makecell[l]{T. Wang et al.} & \makecell[l]{ResNeSt-101/200~\cite{zhang2020resnest}, \\ SEResNeXt-101~\cite{xie2017aggregated}} & \makecell[l]{HorizontalFlip, FiveCrop, TenCrop,\\ label smoothing~\cite{szegedy2016rethinking}} & \makecell[l]{\textbf{Iterative Partition-based} \\ \textbf{Invariant Risk Minimization}} & 71.6          \\ \midrule
\multicolumn{2}{@{}l}{\textbf{Object detection}} & & & \\ \midrule
1 & \makecell[l]{\textbf{Lu et al.}} & YOLO 4-5~\cite{bochkovskiy2020yolov4, yolo5}  & \makecell[l]{Mosaic~\cite{bochkovskiy2020yolov4},  MixUp~\cite{zhang2018mixup}, \\  random color-jittering} & 
\makecell[l]{Weighted Boxes Fusion~\cite{solovyev2021weighted}} & \textbf{30.5}       \\
1$\,\mathsection$ & \makecell[l]{\textbf{Zhang et al.}} & \makecell[l]{Cascade RCNN~\cite{cai2018cascade}, \\
DCN~\cite{dai2017deformable}} & 
\makecell[l]{Multi-scale augmentation, TTA} & \makecell[l]{MoCo v2~\cite{chen2020improved}, Soft-NMS~\cite{bodla2017soft}, \\ Class-specific IoU thresholds} & \textbf{30.4}           \\
2 & \makecell[l]{Niu et al.} &  Swin-T~\cite{liu2021Swin} & Hierarchical labeling & \makecell[l]{FPN~\cite{lin2017feature}, Soft-NMS~\cite{bodla2017soft},\\ pseudo labeling} & 30.4           \\
2$\,\mathsection$  & \makecell[l]{Luo et al.} & \makecell[l]{Cascade RCNN~\cite{cai2018cascade}, \\
DCN~\cite{dai2017deformable}}& Albu, Top-Bottom Cut &  \makecell[l]{GCNet~\cite{cao2019gcnet},  SimSiam~\cite{chen2021exploring}, \\ Soft-NMS~\cite{bodla2017soft}} & 30.1         \\ 
\midrule
\multicolumn{2}{@{}l}{\textbf{Instance segmentation}} & & & \\ \midrule
1 \& J & \makecell[l]{\textbf{Yunusov et al.}~\cite{yunusov2021instance}} &
\makecell[l]{CBSwin-T~\cite{liang2021cbnetv2}} &
\makecell[l]{\textbf{Location-aware MixUp~\cite{zhang2018mixup}}, \\ RandAugment~\cite{NEURIPS2020_d85b63ef}, \\GridMask~\cite{chen2020gridmask}, \\ Random scaling} &
\makecell[l]{Hybrid Task Cascade~\cite{Chen_2019_CVPR}} &
\textbf{47.7} \\
2 & \makecell[l]{Yan et al.~\cite{yan2021second}} &
\makecell[l]{ResNet-101~\cite{he2015deep}} &
\makecell[l]{Random brightness, color jitter,\\saturation, sharpening, blurring,\\noise, pixel shuffle, pixelization,\\filtering, hue transform} &
\makecell[l]{Cascade R-CNN~\cite{Cai_2018_CVPR} \\ Switchable atrous convs.~\cite{qiao2020detectors} \\ Group normalization~\cite{Wu2018GroupN}}
& 40.2 \\
3 & \makecell[l]{Chen et al.} & \makecell[l]{Swin~\cite{liu2021Swin}} &
\makecell[l]{HorizontalFlip, \\ Random scale and crop} & Cascade Mask-RCNN~\cite{cai2018cascade} &
36.6 \\
4 & \makecell[l]{Chen, Zheng} & ResNet-50~\cite{he2015deep} &
\makecell[l]{Instaboost~\cite{fang2019instaboost}} &
\makecell[l]{SCNet~\cite{vu2019cascade}, Seesaw Loss~\cite{Wang_2021_CVPR}, \\ Deformable Convolutions~\cite{dai2017deformable}} &
18.5 \\ \midrule
\multicolumn{2}{@{}l}{\textbf{Action recognition}} & & & \\ \midrule
1 \& J & \makecell[l]{\textbf{Dave et al.}} & R3D\cite{r3d}, I3D\cite{i3d}, MViT\cite{mvit} &  & TCLR\cite{tclr} & 0.74 \\
1 & \makecell[l]{Wu et al.} & TPN\cite{tpn}, Slowfast (slow path)\cite{Feichtenhofer2019sf} & MixUp~\cite{zhang2018mixup}, CutMix~\cite{yun2019cutmix} & & 0.66 \\
2 & \makecell[l]{Gao et al.} & \makecell[l]{Swin~\cite{swin21}, TPN~\cite{tpn}, X3D\cite{x3d},\\ R2+1D\cite{Tran2018r21d}, TimesFormer\cite{timesformer}, \\ Slowfast\cite{Feichtenhofer2019sf}} & & & 0.73\\ \midrule
\multicolumn{2}{@{}l}{\textbf{Re-identification}} & & & \\
\midrule
1 & \makecell[l]{\textbf{Liu et al.}} & \makecell[l]{ResNet \cite{he2015deep}, ResNetSt \cite{zhang2020resnest},\\SE-ResNetXt \cite{xie2017aggregated}\\(24 models)} & \makecell[l]{Difficult sample mining \cite{Shrivastava_2016_CVPR},\\Random Erasing \cite{zhong2020random},\\Local Grayscale Transform,\\affine transformations,\\pixel padding, random flip.} & \makecell[l]{Triplet loss and circle loss,\\with augmentation test,\\re-ranking \cite{zhong2017re},\\query expansion \cite{chum2007total}} & \textbf{96.5} \\
2 \& J & \makecell[l]{Chen et al.} & \makecell[l]{ResNet-IBN \cite{he2015deep}, \\ SE ResNet-IBN \cite{hu2018squeeze} (5 models)} & \makecell[l]{Video temporal mining,\\Random Erasing \cite{zhong2020random},\\pixel padding, random flip} & \makecell[l]{Cross-entropy and Triplet loss,\\with augmentation test,\\re-ranking \cite{zhong2017re},\\6x schedule \cite{he2019rethinking}.} & 96.4 \\
3 & \makecell[l]{Qi et al.} & \makecell[l]{ResNet-IBN \cite{he2015deep},\\23OSNet on Stronger Baseline}& \makecell[l]{Random Erasing \cite{zhong2020random},\\color jitter, random flip, \\AutoAugment~\cite{cubuk2018autoaugment}} & \makecell[l]{Cross-entropy and Triplet loss,\\re-ranking \cite{zhong2017re},\\query expansion \cite{chum2007total}} & 94.2 \\
4 & \makecell[l]{Zheng et al.} & \makecell[l]{ResNet-IBN \cite{he2015deep}\\w/ spatial and channel attention}& \makecell[l]{Random Erasing and Patch \cite{zhong2020random},\\color jitter, random flip, \\AutoAugment~\cite{cubuk2018autoaugment}} & \makecell[l]{Cross-entropy,\\Triplet loss and circle loss} & 84.8 \\
\bottomrule
\end{tabular}
}
\end{table*}
\section{Challenges}
\label{sec:challenges}

Slightly different from the 1st Visual Inductive Priors for Data-Efficient Deep Learning Workshop~\cite{bruintjes2021vipriors}, this year, the workshop accommodates five common computer vision challenges in which  the number of training samples are reduced to a small fraction of the full set:

\textbf{Image classification}: We use a subset of Imagenet~\cite{deng2009imagenet}. The subset contains 50 images from 1,000 classes for training, validation and testing.

 \textbf{Object detection}: DelftBikes~\cite{kayhan2021hallucination} dataset is used for this challenge. The dataset includes 8,000 bike images for training  (Fig.~\ref{fig:bike_images}). In each image, 22 different bike parts are annotated with bounding box, class and object state labels.

\textbf{Instance segmentation}: The main objective of the challenge is to segment basketball players and the ball on images recorded of a basketball court. The dataset is provided by SynergySports\footnote{\url{https://synergysports.com}} and contains a train, validation and test set of basketball games recorded at different courts with instance labels.

\textbf{Action recognition}: For this challenge we have provided Kinetics400ViPriors, which is an adaptation of the well-known Kinetics400 dataset~\cite{kinetics400}. The training set consists of approximately 40k clips, while the validation and test sets contain about 10k and 20k clips, respectively.

\textbf{Re-identification}: The dataset for the person re-identification task is provided by Synergy Sports. The training set contains 436 identities (ids) in 8569 images (around 20 images per id). A validation set is also provided with 50 query ids for a gallery of 910 images. Finally, the test set is composed by a query set of 468 ids for a gallery of 8703 images.

We provide a toolkit\footnote{\url{https://github.com/VIPriors/vipriors-challenges-toolkit}} which consists of guidelines, baseline models and datasets for each challenge.
The competitions are hosted on the Codalab platform. Each participating team submits their predictions computed over a test set of samples for which labels are withheld from competitors.

The challenges include certain rules to follow:
\begin{itemize}
    \item Models ought to train from scratch with only the given dataset.
    \item The usage of other data rather than the provided training data, pretraining the models and transfer learning methods are prohibited. 
    \item The participating teams need to write a technical report about their methodology and experiments.
\end{itemize}

\textbf{Shared rankings.} Due to confusion around the exact deadline of the competitions, we have merged rankings of two different moments. This has resulted in shared places in some of the rankings of the individual challenges.
\subsection{Classification}

Image classification serves as an important benchmark for the progress of deep computer vision research. In particular, the ImageNet dataset~\cite{deng2009imagenet} has been the go-to benchmark for image classification research. ImageNet gained popularity because of its significantly larger scale than those of existing benchmarks. Ever since, even larger datasets have been used to improve computer vision, such as the Google-owned JFT-300M~\cite{sun2017revisiting}. However, we anticipate that relying on the increasing scale of datasets is problematic, as increased data collection is expensive and can clash with privacy interests of the subjects. In addition, for domains like medical imaging, the amount of labeled data is limited and the collection and annotation of such data relies on domain expertise. Therefore, we posit that the design of data efficient methods for deep computer vision is crucial. 

As last year, in our image classification challenge we provide a subset~\cite{kayhan2020translation} of the Imagenet dataset~\cite{deng2009imagenet} consisting of 50 images per class for each of the train, validation and test splits. 
The classification challenge had 14 participating teams, of which six teams submitted a report. The final ranking and the results can be seen in Table~\ref{tab:classification}.

\begin{table*}[t]
\centering
\caption{Final rankings of the Image Classification challenge. $\dagger$ indicates ranks achieved at the close of the original deadline (see Sec.~\ref{sec:challenges}).}
\renewcommand{\arraystretch}{2.0}
\begin{tabular}{@{}llc@{}}
\toprule
Ranking & Teams                & Top-1 Accuracy \\ \midrule
1       & \makecell[l]{\textbf{Pengfei Sun, Xuan Jin, Xin He, Huiming Zhang, Yuan He, Hui Xue.} \\ \textbf{\textit{Alibaba Group.}}}   & \textbf{75.5}           \\
2       & \makecell[l]{Jiahao Wang, Hao Wang, Yifei Chen, Yanbiao Ma, Fang Liu, Licheng Jiao. \\ \textit{School of Artificial Intelligence, Xidian University.}} & 75.2           \\
2$\,\dagger$       & \makecell[l]{Yilu Guo, Shicai Yang, Weijie Chen, Liang Ma, Di Xie, Shiliang Pu. \\ \textit{Hikvision Research Institute.}}  & 74.3           \\
3       & \makecell[l]{Tan Wang, Wanqi Yin, Jiaxin Qi, Jin Liu, Jayashree Karlekar, Hanwang Zhang. \\ \textit{Nanyang Technological University and Panasonic R\&D Center Singapore.}} & 71.6          \\ 
4       & \makecell[l]{Björn Barz, Lorenzo Brigato, Luca Iocchi, Joachim Denzler. \\ \textit{Friedrich Schiller University Jena and Sapienza University of Rome.}} & 69.7            \\
5       & \makecell[l]{Xinran Song, Chang Liu, Wenxin He. \\ \textit{Xidian University.}} & 68.6 \\ \midrule
Jury prize       & \makecell[l]{Tan Wang, Wanqi Yin, Jiaxin Qi, Jin Liu, Jayashree Karlekar, Hanwang Zhang. \\ \textit{Nanyang Technological University \& Panasonic R\&D Center Singapore.}} & 71.6 \\
\bottomrule
\end{tabular}
\label{tab:classification}
\end{table*}
\subsubsection{First place}
\begin{figure}[h]
	\centering
	\includegraphics[width=\linewidth]{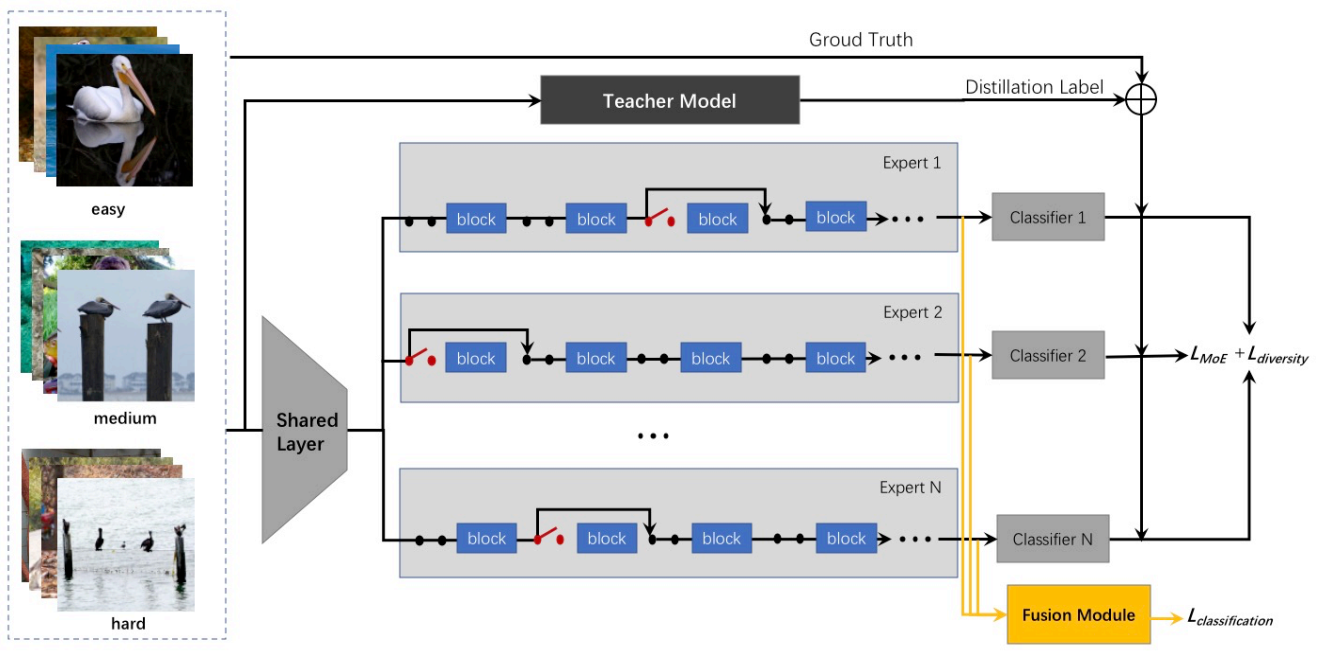}
	\caption{Method used by first place competitor: team Alibaba. Reproduced from internal report with permission.}
	\label{fig:cls-1-first}
\end{figure}

The team from Alibaba Group uses a Mixture of Experts models, which learns multiple different neural architectures in parallel, while sharing some initial backbone layers. The embeddings of all experts are fused, after which a final layer makes the final prediction. During training, each expert has a separate classifier, which optimizes against distilled targets generated by a single separate teacher model. Diversity between experts is stimulated by optimizing the negative KL-divergence between the average output of the experts and the current experts output. See Fig.~\ref{fig:cls-1-first}.

\subsubsection{Second place}

The teams of Xidian University and Hikvision share second place. 

The team from Xidian University use model ensembling with a ResNet \cite{he2015deep}, a TResNet \cite{ridnik2021tresnet}, a Rexnet \cite{han2021rethinking} and an Inception-ResNet \cite{szegedy2017inception}. They also use recent data augmentation methods and label smoothing, in combination with an unpublished method called "dynamic semantic scale balance loss" (DSB).

The team from Hikvision use a representation trained with contrastive regularization, adding in Mean Teacher loss \cite{tarvainen2017mean} and Symmetric Cross Entropy loss \cite{wang2019symmetric}. Additionally, they use model ensembling and aggressive data augmentation.

\subsubsection{Third place \& jury prize}

\begin{figure}[h]
	\centering
	\includegraphics[width=0.8\linewidth]{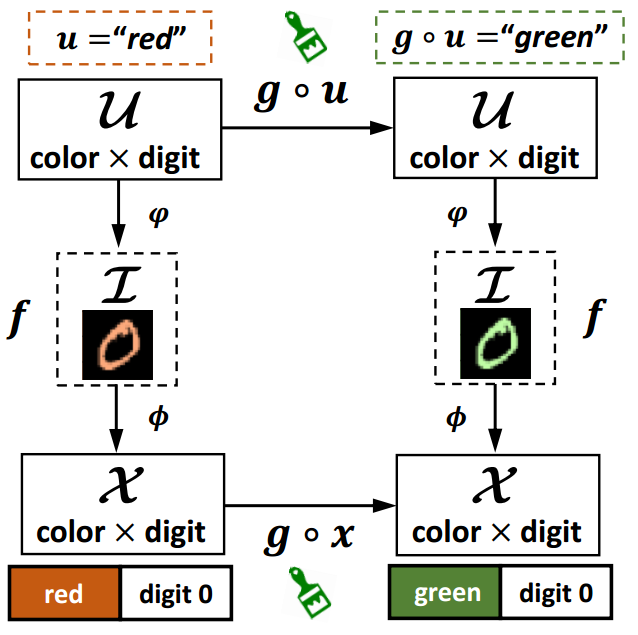}
	\caption{Method proposed by third place competitor \& jury prize winner: team Nanyang \& Panasonic. "Iterative Partition-based Invariant Risk Minimization" (IP-IRM) disentangles semantic concepts in a representation to improve self-supervised learning. Reproduced from internal report with permission.}
	\label{fig:classification-3-disentangling}
\end{figure}

The team from Nanyang Technological University and Panasonic Singapore achieved third place in the competition, as well as the jury prize. Their method first uses self-supervised learning to train a representation, then uses the weights of the learned network to initialize a teacher and a student model in a distillation framework. The distillation network phase is trained with RandAugment \cite{cubuk2020randaugment} and AutoAugment \cite{cubuk2018autoaugment}, label smoothing \cite{pereyra2017regularizing}, random erasing \cite{zhong2020random}.

For the self-supervised learning, the authors propose a novel method called "Iterative Partition-based Invariant Risk Minimization" (IP-IRM), which is grounded in group equivariance, to disentangle the difference semantic concepts of a representation (Fig.~\ref{fig:classification-3-disentangling}).

\subsubsection{Conclusion}

Common patterns (Tab~\ref{tab:conclusion}) in the winning submissions are heavy use of network ensembles and data augmentation. In particular, ResNeSt~\cite{zhang2020resnest} and AutoAugment~\cite{cubuk2018autoaugment} are popular methods to be included in winning submissions. The value of original methodological contributions is hard to estimate, though in this particular competition the jury prize winning submission was able to compete using a significant original contribution and without heavy use of data augmentation.
\subsection{Object Detection}

\begin{figure}[!ht]
	\centering
	\includegraphics[width=\linewidth]{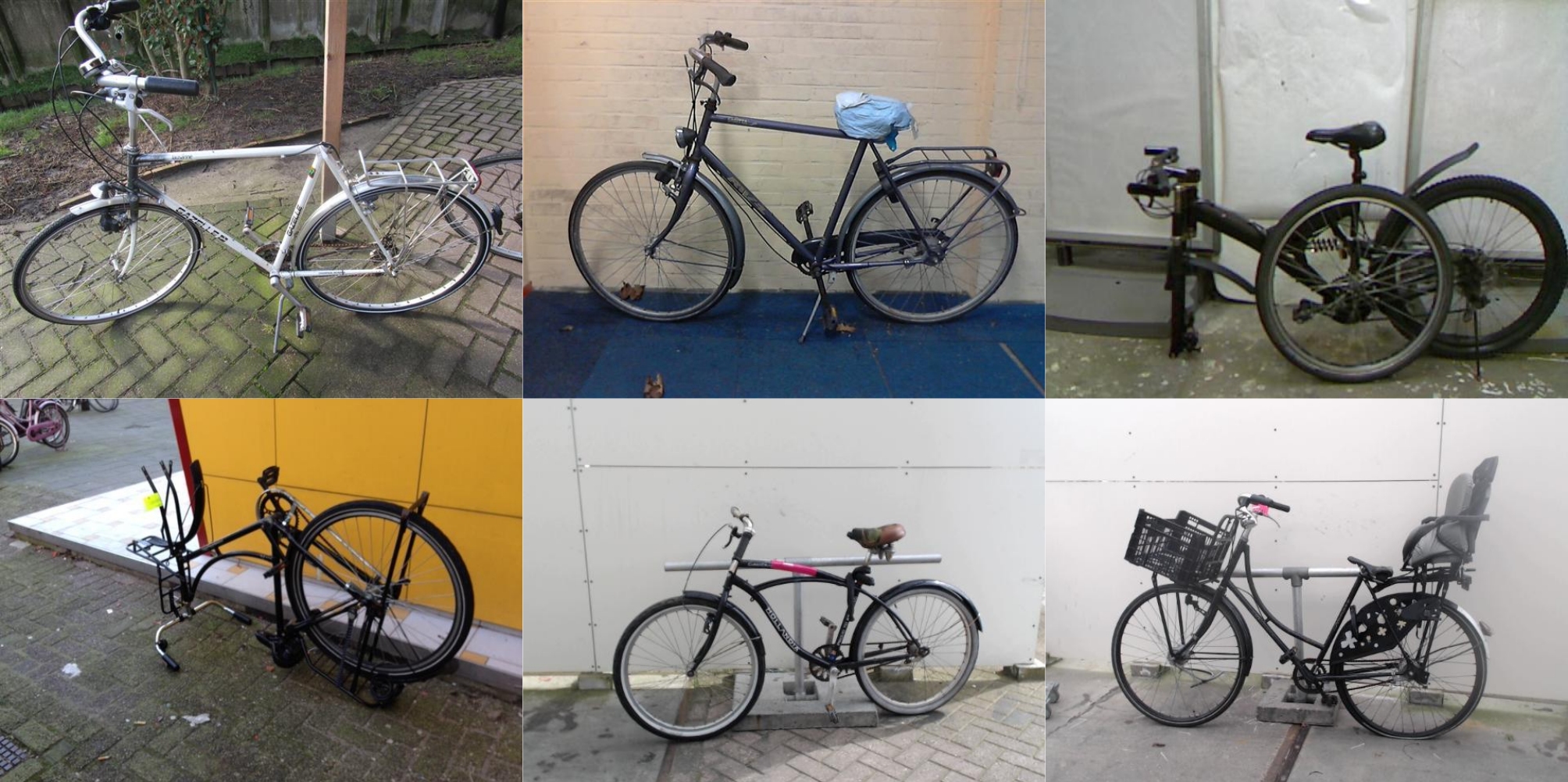}	
	\caption{Example images from DelftBikes dataset. Each image has a single bike with 22 bounding box annotated parts. }
	\label{fig:bike_images}
\end{figure}

In our detection challenge, we use DelftBikes~\cite{kayhan2021hallucination} dataset of 10k bike images (Fig.~\ref{fig:bike_images}). Every image in the dataset includes 22 labeled bike parts such as steer, bell, saddle, front wheel, lamp etc. In addition to the class and bounding box label of each part, the dataset has extra object state labels such as intact, missing, broken or occluded.
The dataset contains varying object sizes, and contextual and location biases which can lead false positive detections. 

As a baseline detector, we train a Faster RCNN detector with a Resnet-50 FPN~\cite{ren2016faster} backbone from scratch for 16 epochs. This baseline method is trained with original image size without any data augmentation and reaches 25.8\% AP score.
To note that, the evaluation is done on available parts, namely intact, damaged and occluded parts.

\begin{table*}[t]
\centering
\caption{Final rankings of the Image Object Detection challenge. $\mathsection$ indicates ranks achieved at the close of the original deadline (see Sec.~\ref{sec:challenges}).}
\renewcommand{\arraystretch}{2.0}
\begin{tabular}{@{}llc@{}}
\toprule
Ranking & Teams                & AP @ 0.5:0.95 \\ \midrule
1       & \makecell[l]{\textbf{Xiaoqiang Lu, Guojin Cao, Xinyu Liu, Zixiao Zhang, Yuting Yang.} \\ \textbf{\textit{School of Artificial Intelligence, Xidian University.}}}   & \textbf{30.5}           \\
1$\,\mathsection$        & \makecell[l]{\textbf{Huiming Zhang, Xuan Jin, Pengfei Sun, Yuan He, Hui Xue.} \\ \textbf{\textit{Alibaba Group.}}} & \textbf{30.4}           \\
2      & \makecell[l]{Junhao Niu, Yu Gu, Luyao Nie, Chao You. \\ \textit{Xidian University.}}  & 30.4           \\
2$\,\mathsection$        & \makecell[l]{ Linfeng Luo, Yanhong Liu, Fengming Cao. \\ \textit{Pingan International Smart City.}} & 30.1         \\ 
 \midrule
Jury prize       & \makecell[l]{Zhang Yuqi.  \\ \textit{ Pingan International Smart City.}} & 43.9 \\
\bottomrule
\end{tabular}
\label{tab:detection}
\end{table*}

The detection challenge had 33 participant teams. The teams from Xidian University and Alibaba Group shared the first place by respectively 30.5\% and 30.4\% AP scores. 


\subsubsection{First places}
\begin{figure}[!ht]
	\centering
	\includegraphics[width=\linewidth]{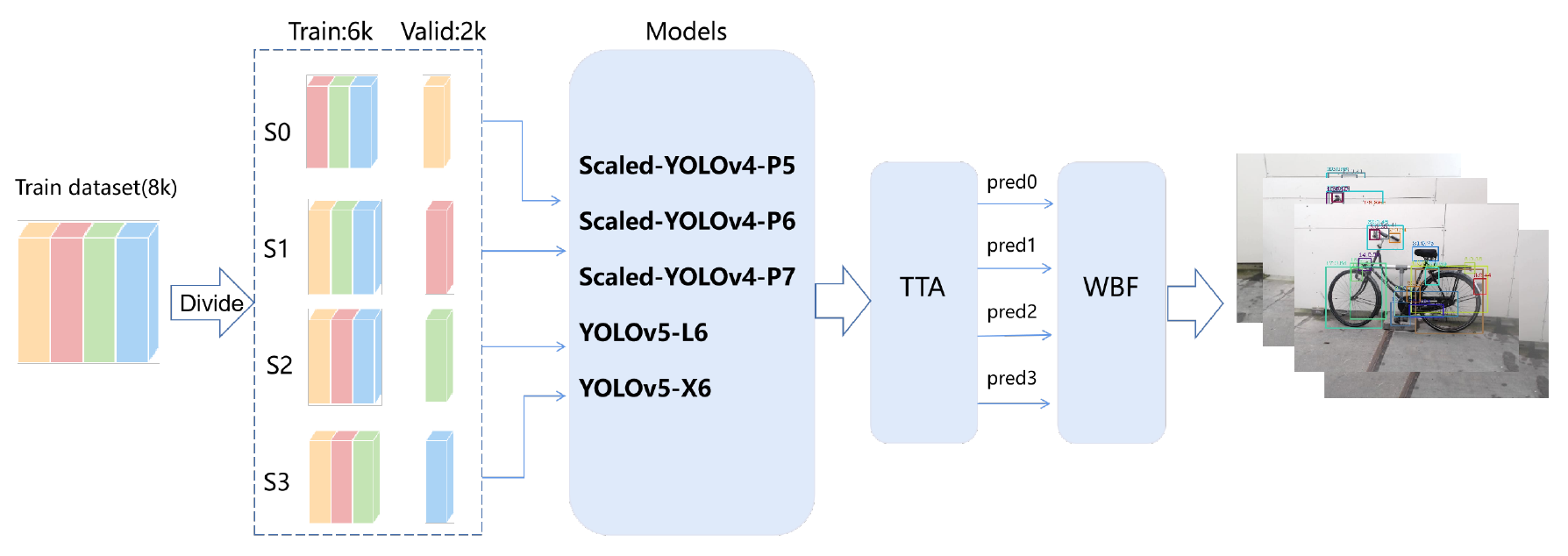}	
	\caption{Multi-scale bagging YOLO detector~\cite{lu2021bagging}. They train different YOLO models and apply weights box fusion (WBF) to obtain better detection.}
	\label{fig:arch11}
\end{figure}

Lu et al.~\cite{lu2021bagging} employ a multi-scale bagging method with various YOLO detectors~\cite{bochkovskiy2020yolov4, yolo5} (Fig.~\ref{fig:arch11}). First, they split the given dataset into 4 independent train and validation sets. Afterwards, they utilize data augmentation methods such as mosaic~\cite{bochkovskiy2020yolov4},  mix-up~\cite{zhang2018mixup} and  random color-jittering. Lastly, weighted boxes fusion (WBF)~\cite{solovyev2021weighted} technique is used to refine predicted boxes from different detectors. The ensemble of 200 models reaches 30.5\% mAP score.

Zhang et. al.~\cite{HuimingCS21} use Cascade RCNN~\cite{cai2018cascade} with ResNet-50~\cite{he2015deep} backbone with DCN~\cite{dai2017deformable}. First, they start with self-supervised learning method MoCo~\cite{chen2020improved} to pretrain the model. Afterwards, they train the model for 24 epochs with multi-scale image sizes. Instead of NMS, they use Soft-NMS~\cite{bodla2017soft} and category-specific IoU thresholds.
In addition, box ensemble of Cascade RCNN and Double-Head (DH) Faster RCNN~\cite{ren2016faster} is applied. The Cascade RCNN detector obtains better result on small objects, yet the DH Faster RCNN does better on larger objects. The authors also use test time augmentation to reach their final AP score of 30.4\%.

\subsubsection{Second places}
\begin{figure}[!ht]
	\centering
	\includegraphics[width=\linewidth]{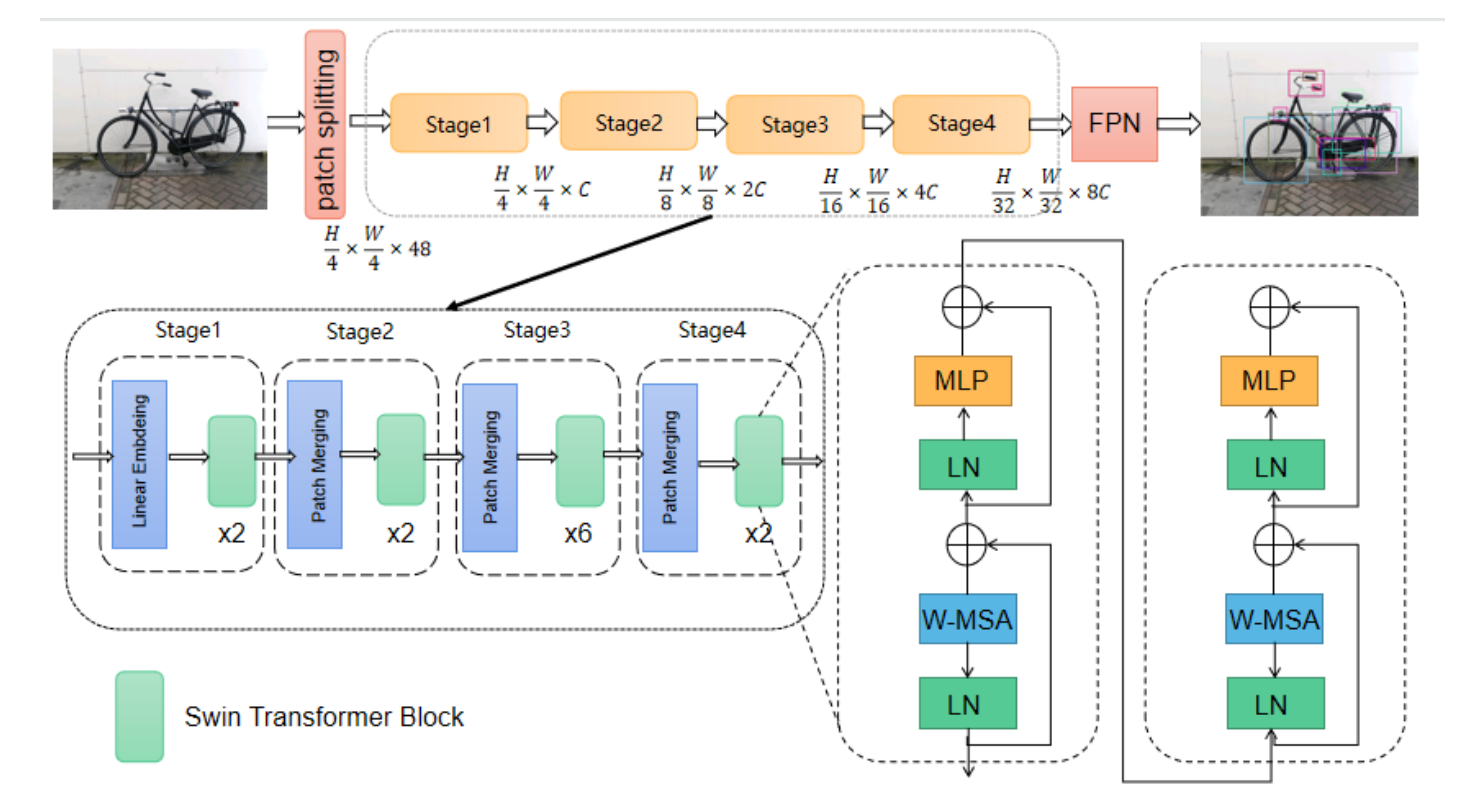}	
	\caption{Swin Transformer with FPN method~\cite{swin21}. Swin-T model inputs 4x4 image patches. FPN follows Swin Transformer blocks.}
	\label{fig:swindet}
\end{figure}

Niu et al.~\cite{swin21} train Swin Transformer~\cite{liu2021Swin} with 4x4 patch inputs. Swin-T Transformer architecture with 96 channels is followed by Feature Pyramid Network (FPN)~\cite{lin2017feature} (Fig.~\ref{fig:swindet}). 
They train the detector with 6 different image sizes with a multi-scale manner for 50 epochs. They also train Deformable DETR Transformer~\cite{zhu2020deformable}, yet Swin Transformer outperforms on DelftBikes~\cite{kayhan2021hallucination} dataset by 2\%. Applying Soft-NMS~\cite{bodla2017soft}, pseudo labeling and dividing the part classes in 2 parts increase both Deformable DETR and Swin Transformers results respectively 29.9\% and 30.4\%.

Luo et al.~\cite{Linfeng21} use Cascade RCNN~\cite{cai2018cascade} coupled with Deformable Convolution Network (DCN)~\cite{dai2017deformable}  and Global Context Modeling Network (GCNet)~\cite{cao2019gcnet}. First, they create synthetic dataset with 10K images for contrastive self-supervised learning by using SimSiam~\cite{chen2021exploring} method. In addition, they utilize multi-scale training and testing, data augmentation and Soft-NMS~\cite{bodla2017soft}. With all the tricks, the method obtains 30.1\% AP performance on the test set.
\subsubsection{Jury prize}
Zhang ~\cite{nan2021} has a very interesting approach, just filling the bounding box locations with \textit{NaN}, the author obtains the highest score in the challenge as 43.9\% mAP. Namely, the author filled the bouding box location of first 10 classes as [NaN, NaN, NaN, NaN]. When giving NaN locations for every classes on validation set, the score becomes 97\%. Therefore, the author takes jury prize of object detection challenge since the author found the bug in Coco API.

\subsection{Instance Segmentation}
Instance segmentation, i.e. the task of detecting and segmenting specific objects and instances in an image is a key problem in computer vision with applications ranging from autonomous driving, surveillance, remote sensing and sport analysis. Our challenge is based on a basketball dataset consisting of images recorded during various basketball games played on different courts, and contain instance segmentation labels for the players and the ball. The train, validation and test splits contain 184, 62, and 64 samples, respectively. The dataset is provided by SynergySports and the test labels are withheld from the challenge participants. The instance segmentation predictions are evaluated by the Average Precision (AP) @ 0.50:0.95 metric. Our baseline method is based on the Detectron2\cite{wu2019detectron2} implementation of Mask-RCNN\cite{}.

Twelve teams submitted solutions to the evaluation server, of which four teams submitted a report to qualify their submission to the challenge. The final rankings are shown in Table \ref{tab:segmentation}.

\begin{table*}[t]
\centering
\caption{Final rankings of the Instance Segmentation challenge.}
\renewcommand{\arraystretch}{2.0}
\begin{tabular}{@{}llc@{}}
\toprule
Ranking & Teams                & \% AP @ 0.50:0.95 \\ \midrule
1       & \makecell[l]{\textbf{Jahongir Yunusov, Shohruh Rakhmatov, Abdulaziz Namozov, Abdulaziz Gaybulayev, Tae-Hyong Kim.} \\ \textbf{\textit{Department of Computer Engineering, Kumoh National Institute of Technology.}}}   & \textbf{47.7}           \\
2       & \makecell[l]{Bo Yan, Fengliang Qi, Leilei Cao, Hongbin Wang. \\ \textit{Ant Group.}} & 40.2           \\
3       & \makecell[l]{Pengyu Chen, Wanhua Li, Jiwen Lu. \\ \textit{Department of Automation, Tsinghua University \& Beijing University of Posts and Telecommunications.}} & 36.6          \\ 
4       & \makecell[l]{Zhenhong Chen, Ximin Zheng.} & 18.5 \\ \midrule
Jury prize       & \makecell[l]{Jahongir Yunusov, Shohruh Rakhmatov, Abdulaziz Namozov, Abdulaziz Gaybulayev, Tae-Hyong Kim. \\ \textit{Department of Computer Engineering, Kumoh National Institute of Technology.}} & 0.477 \\
\bottomrule
\end{tabular}
\label{tab:segmentation}
\end{table*}

\subsubsection{First place \& jury prize}

\begin{figure}
    \centering
    \includegraphics[width=\linewidth]{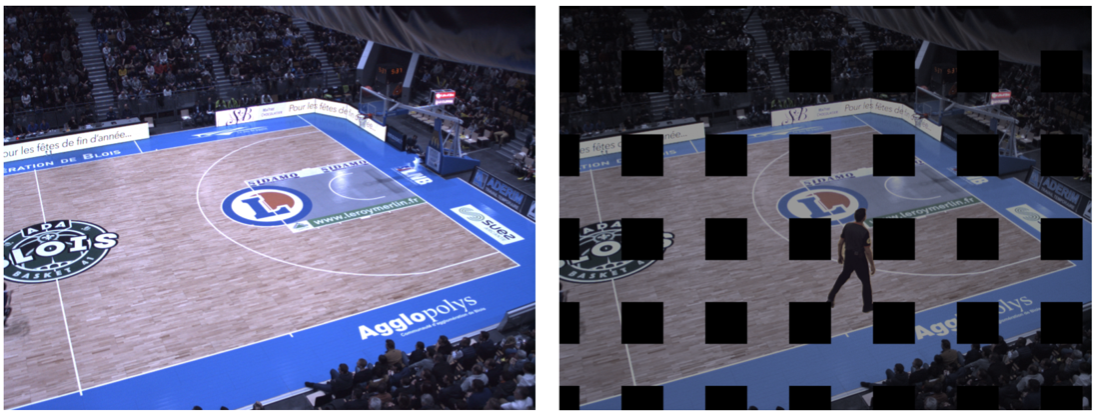}
    \caption{Data augmentation policy of the first-place instance segmentation submission by Yunusov et al.~\cite{yunusov2021instance}. Original image shown left, augmented image shown right.}
    \label{fig:3_1_seg}
\end{figure}

The method of Yunusov et al.~\cite{yunusov2021instance} is based on the HTC detector~\cite{Chen_2019_CVPR} and the CBSwin-T backbone with CBFPN~\cite{liang2021cbnetv2} using group normalization. During training, the multi-scale sampling mode from~\cite{mmdetection} is used. Inference is performed on a single fixed scale. The authors propose a simple yet effective data augmentation scheme based on~\cite{zhang2018mixup}, where instances are cropped based on their segmentation masks and are copied onto different images, while maintaining class balance. The locations where the instances are placed are constrained to be realistic, i.e. to always lie within the basketball court. Additionally, the RandAugment~\cite{NEURIPS2020_d85b63ef} and GridMask~\cite{chen2020gridmask} data augmentation methods are employed. The complete data augmentation policy is illustrated in figure~\ref{fig:3_1_seg}.

Due to the simplicity and effectiveness of the method, this submission has additionally been awarded the jury prize.

\subsubsection{Second place}

\begin{figure}
    \centering
    \includegraphics[width=\linewidth]{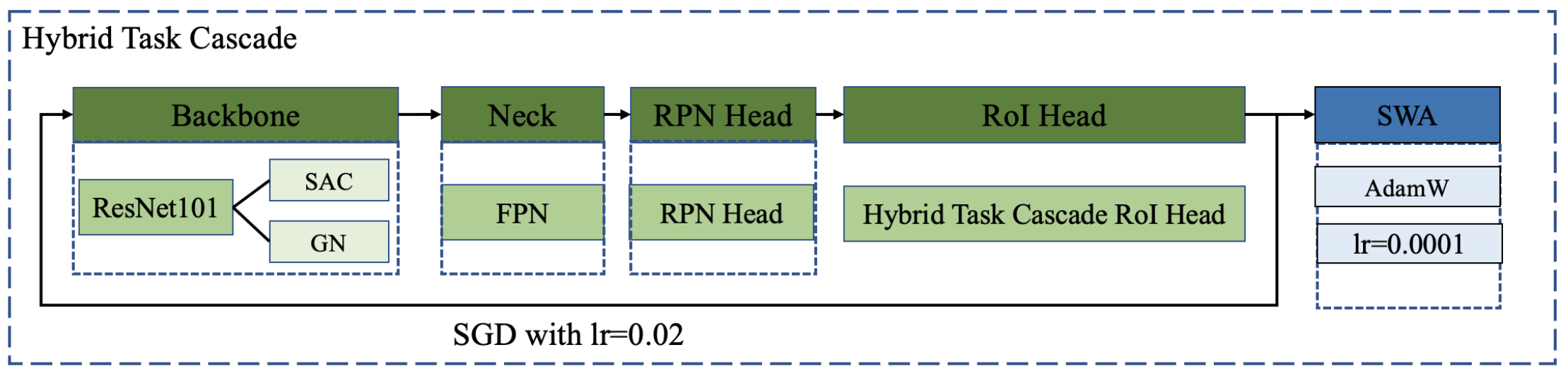}
    \caption{Instance segmentation model architecture and training pipeline of the method of Yan et al.}
    \label{fig:3_2_seg}
\end{figure}

Yan et al.~\cite{yan2021second} propose a combination of offline and online augmentation, where the dataset is first expanded by generating ten augmented versions for each original image. Offline augmentations include color transformations (random brightness, color jitter, saturation, and sharpen), quality transformations (random blur, noise, pixel shufflin, and pixelization), filter transformations from the PIL.ImageFilter library~\cite{clark2015pillow}, and hue transformations. Random flip, random cropping, bbox-jitter and grid-mask~\cite{chen2020gridmask} are performed during online augmentation. The segmentation model is based on Hybrid Task Cascade (HTC)~\cite{Chen_2019_CVPR}, using a ResNet-101~\cite{he2015deep} with switchable atrous convolutions~\cite{detectors} and group normalization~\cite{Wu2018GroupN}. The model architecture and training pipeline are snown in figure~\ref{fig:3_2_seg}. The model is trained on a single GPU.

\subsubsection{Third place}

\begin{figure}
    \centering
    \includegraphics[width=\linewidth]{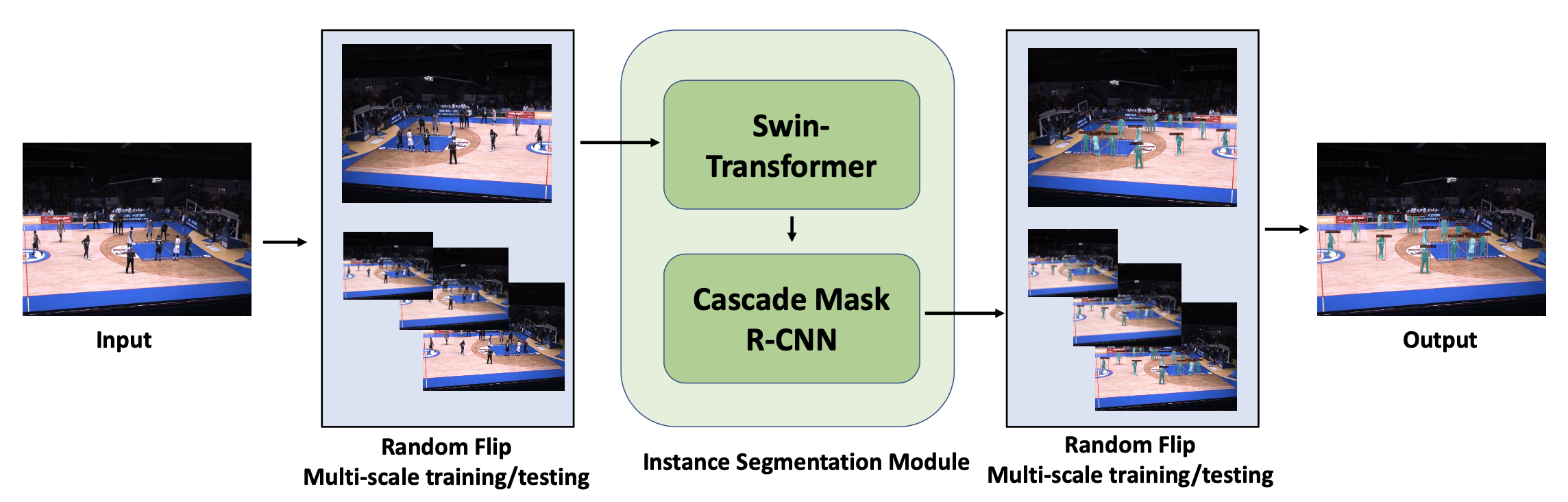}
    \caption{Overview of instance segmentation method by the third place competitor, Chen et al.}
    \label{fig:3_3_seg}
\end{figure}

The method of Chen et al. is based on Cascade R-CNN~\cite{Cai_2018_CVPR}. The Swin Transformer~\cite{liu2021Swin} is used as the feature extractor. During training, a similar data augmentation strategy as in~\cite{liu2021Swin} is used, which includes random flips, scaling and cropping. At test time, multi-scale fusion is performed to improve performance. The model is trained on a single GPU. The overall framework is depicted in figure~\ref{fig:3_3_seg}.

\subsection{Action Recognition}

Many of the most popular Action Recognition models consist of very deep networks whose training process requires a massive amount of data, in the form of frames or clips. This fact becomes one of the main obstacles on occasions when, for example, there is not enough data available or resources are insufficient to be able to adjust the model correctly.

Following the spirit of the workshop, we have provided the Kinetics400ViPriors dataset, which is an adaptation of the well-known Kinetics400 \cite{kinetics400} dataset. We have built a reduced version with only 40k, 10k and 20k clips for the train, validation and test sets, but kept the original number of action classes. With this, we want to encourage Action Recognition researchers to develop efficient models capable of extracting visual prior knowledge from data.

As metric, we evaluate the average classification accuracy over all classes on the test set. The accuracy for one class is defined as $ \mathrm{Acc} = \frac{P}{N} $, where P corresponds to the number of correct predictions for the class being evaluated and N to the total number of samples of the class. The average accuracy is the average of accuracies over all classes.

\subsubsection{Final Rankings}
9 teams submitted solutions to the evaluation server, of which 3 teams submitted a report to qualify their submission to the challenge. The final rankings are shown in Table \ref{tab:actionrecognition}.

\begin{table*}[t]
\centering
\caption{Final rankings of the Action Recognition challenge. $\mathsection$ indicates ranks achieved at the close of the original deadline (see Sec.~\ref{sec:challenges}).}
\renewcommand{\arraystretch}{2.0}
\begin{tabular}{@{}llc@{}}
\toprule
Ranking & Teams                & Acc \\ \midrule
1       & \makecell[l]{\textbf{Ishan Dave, Naman Biyani, Brandon Clark, Rohit Gupta, Yogesh Rawat and Mubarak Shah} \\ \textbf{\textit{Center for Research in Computer Vision (CRCV), University of Central Florida.}}}   & \textbf{0.74}           \\
1$\,\mathsection$       & \makecell[l]{\textbf{Jie Wu, Yuxi Ren and Xuefen Xiao} \\ \textbf{\textit{ByteDance Inc.}}} & \textbf{0.66}           \\
2      & \makecell[l]{Zihan Gao, Tianzhi Ma, Jiaxuan Zhao, Lichen Jiao and Fang Liu \\ \textit{Xidian University.}}  & 0.73   \\ 
 \midrule
Jury prize       & \makecell[l]{\textbf{Ishan Dave, Naman Biyani, Brandon Clark, Rohit Gupta, Yogesh Rawat and Mubarak Shah} \\ \textbf{\textit{Center for Research in Computer Vision (CRCV), University of Central Florida.}}}   & \textbf{0.74}           \\
\bottomrule
\end{tabular}
\label{tab:actionrecognition}
\end{table*}

\subsubsection{First places \& jury prize}
One of the first places and the jury prize of the Action Recognition challenge goes to Dave et al. from CRCV. To tackle the challenge of obtaining a model with the condition of not using any pre-training, Dave et al. proposed a combination of several state-of-the-art techniques that have shown good results in similar circumstances. They decided to use both convolutional (R3D\cite{r3d} and I3D\cite{i3d}) and attention-based (MViT\cite{mvit}) models. As for the convolutional ones, the self-supervised training process TCLR\cite{tclr} is applied first. Then, the resulting models are finetuned using RGB and optical flow frames. On the other hand, the transformer model MViT is trained directly on Kinetics400ViPriors using only RGB frames.

\begin{figure}
    \centering
    \includegraphics[width=\linewidth]{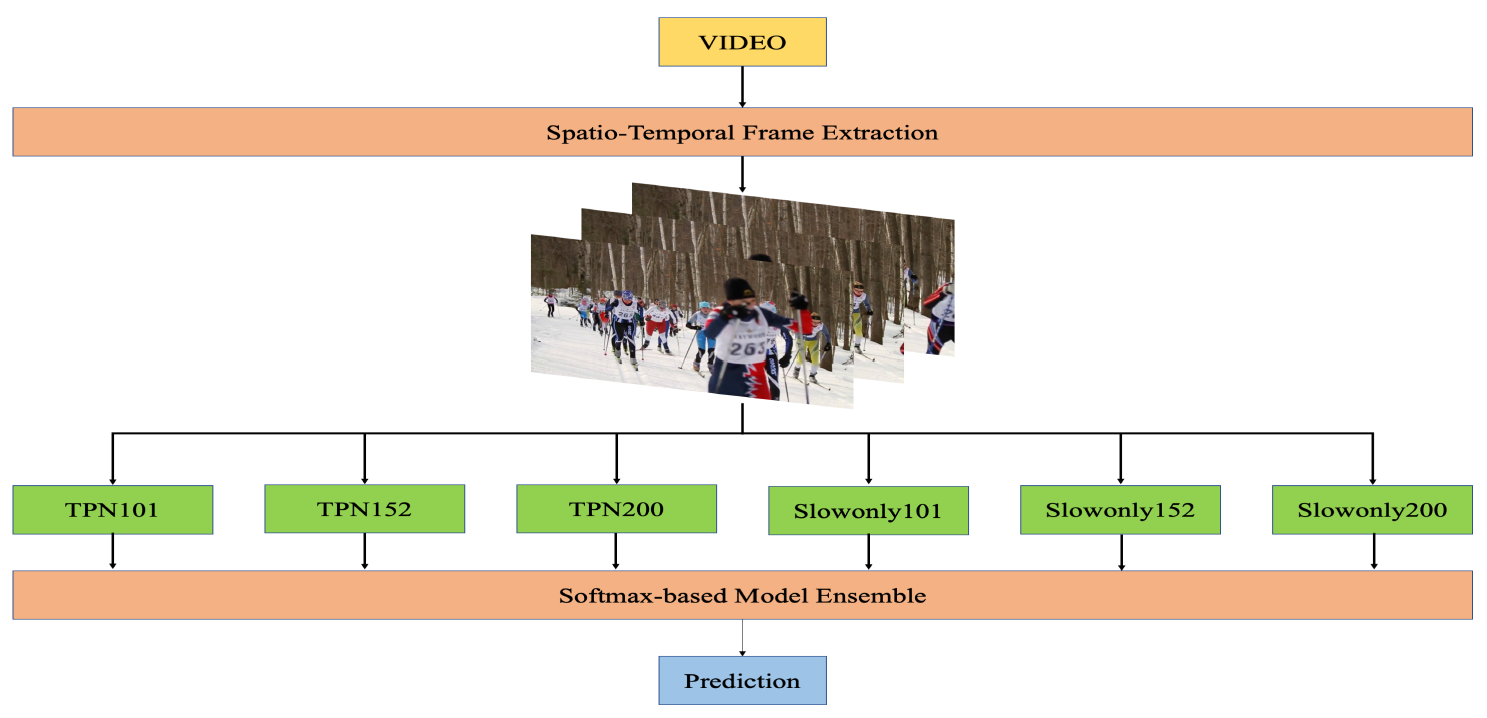}
    \caption{Method proposed by first place Wu et al. from ByteDance Inc.}
    \label{fig:wuetal}
\end{figure}

Dave et al. share the first place with Wu et al. from ByteDance Inc, who proposed the ensemble method shown in Figure \ref{fig:wuetal}. Wu et al. considered that for Action Recognition the visual tempo (or dynamics) of an action plays a vital role. To capture this information, they proposed to fuse three TPN\cite{tpn} architectures. Each of the TPN modules is complemented by a Slowonly network, which is the slow path of the Slowfast network. Additionally, mixup and cutmix data augmentation techniques are applied during training.

\subsubsection{Second place}
\begin{figure}
    \centering
    \includegraphics[width=\linewidth]{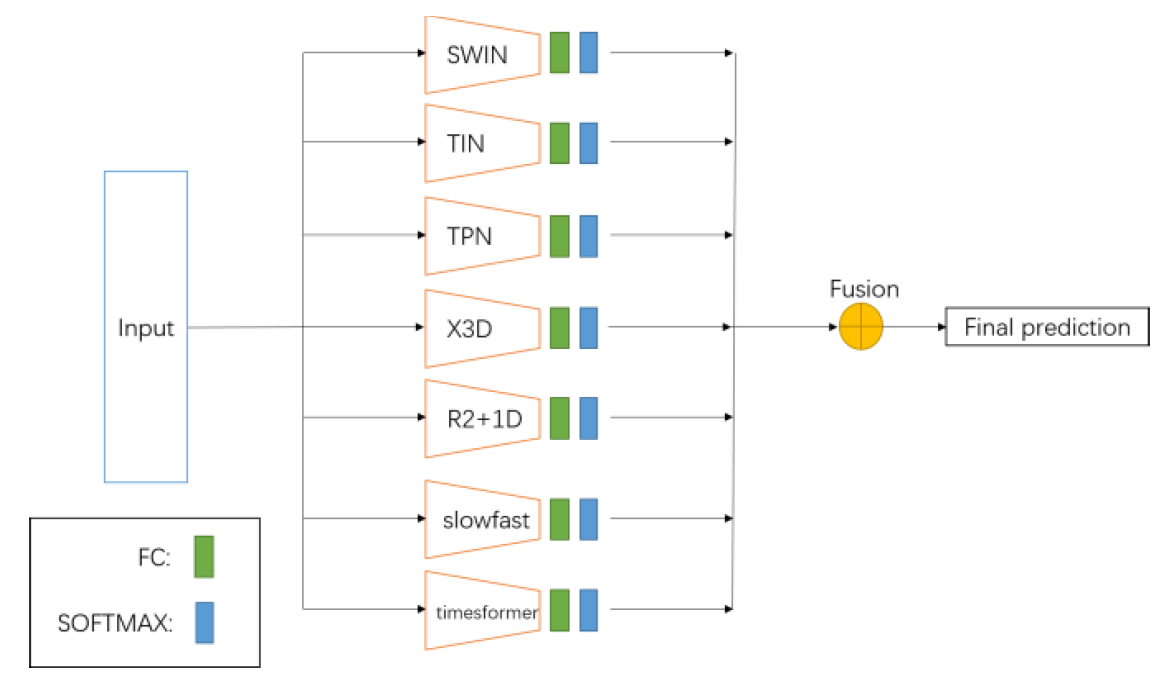}
    \caption{Method proposed by first place Wu et al. from ByteDance Inc.}
    \label{fig:gaoetal}
\end{figure}

The second best ranked team is formed by Gao et al. from Xidian University. As shown in Figure \ref{fig:gaoetal}, their solution is basically a combination of some of the best performing methods for Action Recognition: Swin Transformer\cite{swin21}, TPN\cite{tpn}, X3D\cite{x3d}, R2+1D\cite{Tran2018r21d}, Slowfast\cite{Feichtenhofer2019sf} and Timesformer\cite{timesformer}.

\subsection{Re-identification}
\begin{figure*}[t]
	\centering
	\includegraphics[width=1.0\linewidth]{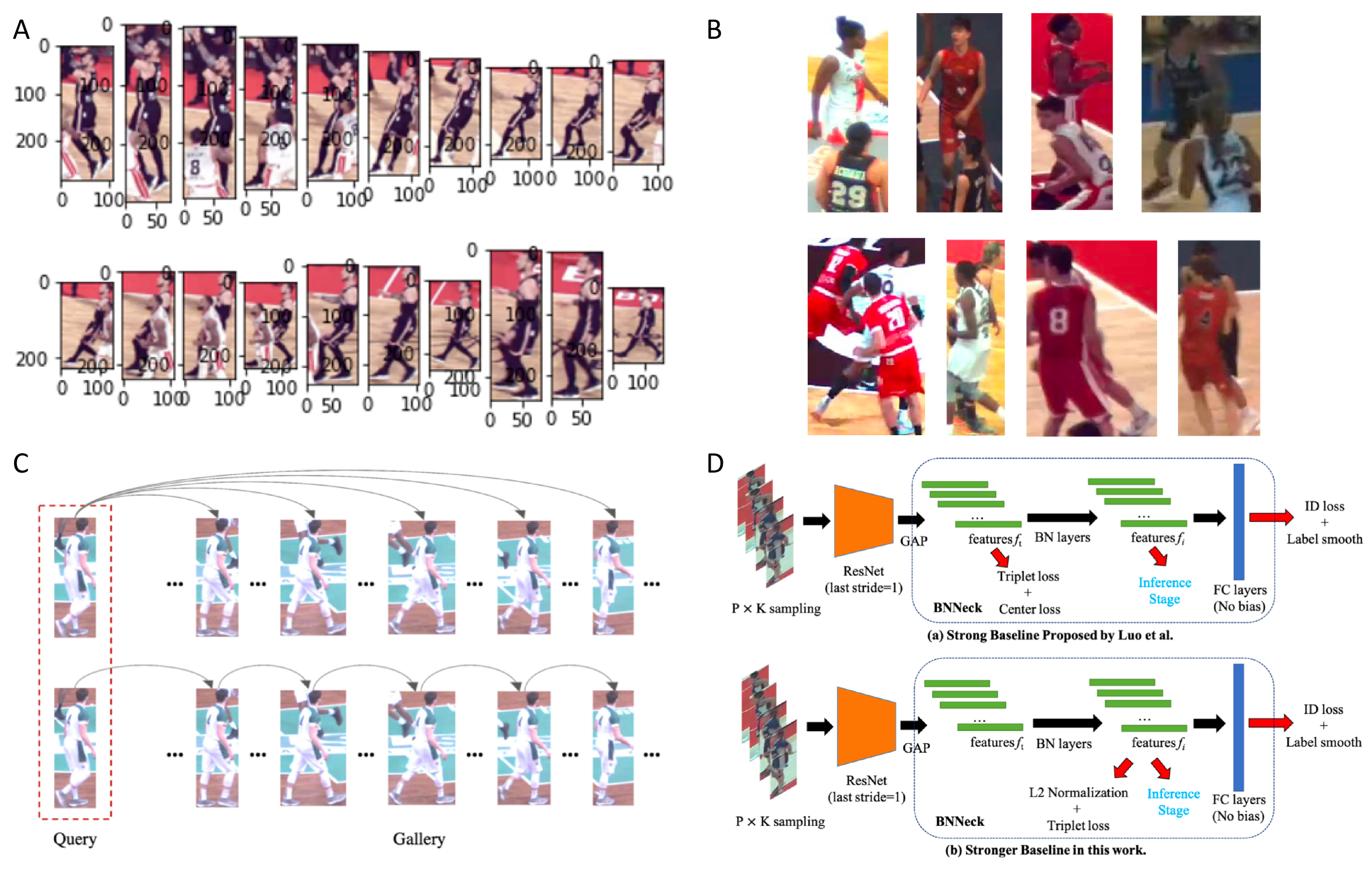}
	\caption{Re-identification challenge summary. A. A sample of 20 images for a single identity in the Synergy Sports Re-id dataset. B. First Place strategy focused on partial (top row) and full (bottom row) occlusions. C. Second Place proposed a post-processing strategy named video temporal relationship mining. D. Third Place proposed a stronger baseline to solve conflicts between cosine metric and euclidean metric spaces.}
	\label{fig:reidsummary}
\end{figure*}

Person re-identification has become an established field in Computer Vision. The idea is to train a model to extract an embedding vector that unequivocally identifies a person identity. Simplifying, the model is usually trained as a classifier and the second-last layer is used as the embedding vector that discriminates people features. During testing, a query identity--different from the identities used for training--has to be found into the gallery: an  embedding vector is extracted from the query image and then compared with all the embedding vectors from the gallery. If two vectors are similar enough, the two images are considered to belong to the same person. Consequently, the more identities the model has been trained to classify during the training phase, the better the embedding vector will be able to extract useful and discriminating information from an unseen person during testing. A well known public dataset for person re-identification, Market-1501 \cite{zheng2015scalable}, contains 1501 identities (700 for training and 701 for testing) and over 32000 annotated bounding boxes. As a comparison, the provided dataset contains 954 identities and 18232 images.
In the person re-identification challenge, the dataset comes from short sequences of basketball games, each sequence is composed by 20 frames. For the validation and test sets, the query images are persons taken at the first frame, while the gallery images are identities taken from the 2nd to the last frame.
Figure \ref{fig:reidsummary} shows an example of one identity in the dataset (\ref{fig:reidsummary} A) and summarizes the contributions of the first 3 submissions: the first proposing a solution for occlusions, the second leveraging the time correlations between frames, and the third improving over the most common architecture for re-identification problems.

\subsubsection{Final Rankings}
Twelve teams submitted solutions to the evaluation server, of which four teams submitted a report to qualify their submission to the challenge. The final rankings are shown in Table \ref{tab:personreidentification}.

\begin{table*}[t]
\centering
\caption{Final rankings of the Person Re-identification challenge.}
\renewcommand{\arraystretch}{2.0}
\begin{tabular}{@{}llc@{}}
\toprule
Ranking & Teams                & Top-1 Accuracy (\%) \\ \midrule
1       & \makecell[l]{\textbf{Cen Liu, Yunbo Peng, Yue Lin} \\ \textbf{\textit{NetEase Games AI Lab}}}   & \textbf{96.5} \\
2       & \makecell[l]{Siyu Chen, Dengjie Li, Lishuai Gao, Fan Liang, Wei Zhang, Lin Ma\\ \textit{Fudan University and Meituan}} & 96.4 \\
3       & \makecell[l]{Fengliang Qi, Bo Yan, Leilei Cao, Hongbin Wang\\ \textit{Ant group}} & 94.0          \\ 
4       & \makecell[l]{XiMin Zheng, JiaQi Yang} & 85.0           \\
Jury prize       & \makecell[l]{Siyu Chen, Dengjie Li, Lishuai Gao, Fan Liang, Wei Zhang, Lin Ma\\ \textit{Fudan University and Meituan}} & 96.4 \\
\bottomrule
\end{tabular}
\label{tab:personreidentification}
\end{table*}

\subsubsection{First place}
The first ranked team based its approach on three components: pre-processing, strong backbones and post-processing.
On pre-processing, the authors started by addressing noisy labels. They applied an online difficult sample mining algorithm similar to \cite{Shrivastava_2016_CVPR} in order to filter out the hard occluded annotations. The occluded annotations were divided into partial occlusions and full occlusions (Figure \ref{fig:reidsummary} B). The latter were removed, while they applied data augmentation on the former to increase their proportion in the training set. This effectively increased the model robustness to occlusions. Moreover, they applied Random Erasing \cite{zhong2020random} and Local Grayscale Transform (LGT) as data augmentation methods. Specifically, LGT avoided color similarities in the jersey. In addition, standard augmentation techniques were adopted such as: affine transformations, pixel padding, random flip. Finally, they oversampled IDs with less than 20 images in gallery to balance the training set.
Concerning the strong backbones, the participants adopted the re-identification baseline introduced in \cite{luo2019bag} together with an ensemble of ResNet \cite{he2015deep}, ResNetSt \cite{zhang2020resnest}, SE-ResNetXt \cite{xie2017aggregated}: a total of 24 models where used in the ensemble, including ResNet-101, ResNet-152, ResNet-200, ResNeSt-101, ResNeSt-152, ResNeSt-200, SE-ResNeX-t101, SE-ResNeXt-152, SE-ResNeXt-200. In addition they applied the Generalized Mean pooling \cite{radenovic2018fine} instead of GAP. Finally, they used a combination of Triplet loss and circle loss. Finally, for post-processing, they applied common techniques for re-id tasks such as: augmentation test, re-ranking \cite{zhong2017re}, query expansion \cite{chum2007total}. 

\subsubsection{Second place \& jury prize}
The second place was nominated as jury price since they managed to extrapolate information from the dataset as a prior and use it within the assignment algorithm obtaining a considerable improvement. They proposed a post-processing strategy named video temporal relationship mining: the first frame of the video (query) is used to retrieve the second frame (in the gallery). Then the second frame is used to retrieve the next frame and so on (see Figure \ref{fig:reidsummary} C, bottom row). This strategy exploited the temporal relationship of the provided data.
The baseline model was the MGN (multiple granularity network, \cite{wang2018learning}): a global branch for global feature representations and local branches for horizontal splits. They observed that using 4 branches (3 split lines) works best for basketball players.
The participants trained an ensemble of models: ResNet-50 with IBN \cite{pan2018two}, while Batch Normalization (BN)\cite{IoffeS15} was replaced by SyncBN with cross-entropy and triple loss. Other common choices came from BoT \cite{luo2019bag} including data augmentation methods such as: random erase \cite{zhong2020random}, random horizontal flipping and padding. As post-processing they used re-ranking \cite{zhong2017re} and 6x Schedule \cite{he2019rethinking}.

\subsubsection{Third place}
The third placed team proposed a Stronger Baseline: a slight modification of Strong baseline \cite{luo2019bag}, providing tiny overhead but faster convergence rate and recognition performance. They see BNNeck as a standardization procedure: Stronger Baseline allows to improve optimization conflicts between Cosine Metric Space and Euclidean metric space. It does so by applying Batch Normalization and L2 Normalization on the last layer of the BNNeck. Figure (\ref{fig:reidsummary} D, bottom row) shows the proposed modifications to the architecture. Interestingly, the authors proved their method on the Market-1501 \cite{zheng2015scalable} dataset as well.
They tested Stronger baseline with OSNet with common data augmentation tools: Horizontal Flip, Random Erasing \cite{zhong2020random}, Color Jitter, AutoAugmentation \cite{cubuk2018autoaugment}; as well as Post Processing strategies: Query Expansion \cite{chum2007total} and re-ranking \cite{zhong2017re}.


\section{Conclusion}

The challenges of the second Visual Inductive Priors for Data-Efficient Deep Learning workshop have yet again given a valuable insight in the current state-of-the-art methods and practices in low-data computer vision. We have summarized all solutions in Table~\ref{tab:conclusion} in terms of the encoder architecture, data augmentation techniques and main methods used. Similarly to the last edition, most participants made heavy use of data augmentation to improve the performance, most importantly AutoAugment~\cite{cubuk2018autoaugment}, MixUp~\cite{zhang2018mixup}, CutMix~\cite{yun2019cutmix} and Random Erasing~\cite{zhong2020random}, besides default geometric and photometric augmentations included in deep learning libraries. Generating synthetic samples proves to be an effective way to increase the training set size and potentially reduce the class imbalance at little to no extra cost. Besides the well-established ResNet architecture several new encoder architectures have been employed, including the Swin transformer~\cite{liu2021Swin, liang2021cbnetv2} and the ResNeSt~\cite{zhang2020resnest} architecture. Interestingly, novel contributions were mainly introduced in the classification challenge, possibly since this is the most fundamental of all problem settings, allowing new ideas to be tested relatively easily. Nevertheless, we encourage next year's participants to experiment with novel ideas in all problem settings, as the additional relevance of absolute spatial location and temporal information allows for more interesting, task-specific priors to be included.

\Urlmuskip=0mu plus 1mu\relax
\bibliographystyle{splncs04}
\bibliography{main}

%




\end{document}